# Efficiency, Robustness and Accuracy in $\mathcal{P}$icky Chart Parsing*

David M. Magerman
Stanford University
Stanford, CA 94305
magerman@cs.stanford.edu

Carl Weir
Paramax Systems
Paoli, PA 19301
weir@prc.unisys.com

## ABSTRACT
This paper describes $\mathcal{P}$icky, a probabilistic agenda-based chart parsing algorithm which uses a technique called *probabilistic prediction* to predict which grammar rules are likely to lead to an acceptable parse of the input. Using a suboptimal search method, $\mathcal{P}$icky significantly reduces the number of edges produced by CKY-like chart parsing algorithms, while maintaining the robustness of pure bottom-up parsers and the accuracy of existing probabilistic parsers. Experiments using $\mathcal{P}$icky demonstrate how probabilistic modelling can impact upon the efficiency, robustness and accuracy of a parser.

## 1. Introduction

This paper addresses the question: Why should we use probabilistic models in natural language understanding? There are many answers to this question, only a few of which are regularly addressed in the literature.

The first and most common answer concerns ambiguity resolution. A probabilistic model provides a clearly defined preference rule for selecting among grammatical alternatives (i.e. the highest probability interpretation is selected). However, this use of probabilistic models assumes that we already have efficient methods for generating the alternatives in the first place. While we have $O(n^3)$ algorithms for determining the grammaticality of a sentence, parsing, as a component of a natural language understanding tool, involves more than simply determining all of the grammatical interpretations of an input. In order for a natural language system to process input efficiently and robustly, it must process all intelligible sentences, grammatical or not, while not significantly reducing the system's efficiency.

This observation suggests two other answers to the central question of this paper. Probabilistic models offer a convenient scoring method for partial interpretations in a well-formed substring table. High probability constituents in the parser's chart can be used to interpret ungrammatical sentences. Probabilistic models can also be used for efficiency by providing a best-first search heuristic to order the parsing agenda.

This paper proposes an agenda-based probabilistic chart parsing algorithm which is both robust and efficient. The algorithm, $\mathcal{P}$icky[1], is considered robust because it will potentially generate all constituents produced by a pure bottom-up parser and rank these constituents by likelihood. The efficiency of the algorithm is achieved through a technique called *probabilistic prediction,* which helps the algorithm avoid worst-case behavior. Probabilistic prediction is a trainable technique for modelling where edges are likely to occur in the chart-parsing process.[2] Once the predicted edges are added to the chart using probabilistic prediction, they are processed in a style similar to agenda-based chart parsing algorithms. By limiting the edges in the chart to those which are predicted by this model, the parser can process a sentence while generating only the most likely constituents given the input.

In this paper, we will present the $\mathcal{P}$icky parsing algorithm, describing both the original features of the parser and those adapted from previous work. Then, we will compare the implementation of $\mathcal{P}$icky with existing probabilistic and non-probabilistic parsers. Finally, we will report the results of experiments exploring how $\mathcal{P}$icky's algorithm copes with the tradeoffs of efficiency, robustness, and accuracy.[3]

## 2. Probabilistic Models in $\mathcal{P}$icky

The probabilistic models used in the implementation of $\mathcal{P}$icky are independent of the algorithm. To facilitate the comparison between the performance of $\mathcal{P}$icky and its predecessor, $\mathcal{P}$earl, the probabilistic model implemented for $\mathcal{P}$icky is similar to $\mathcal{P}$earl's scoring model, the context-free grammar with context-sensitive probability (CFG with CSP) model. This probabilistic model estimates the probability of each parse $T$ given the words in the sentence $S$, $\mathcal{P}(T|S)$, by assuming that each non-terminal and its immediate children are dependent on the non-terminal's siblings and parent and on the part-of-speech trigram centered at the beginning of that rule:

$$\mathcal{P}(T|S) \simeq \prod_{A \in T} \mathcal{P}(A \to \alpha | C \to \beta A \gamma, a_0 a_1 a_2) \quad (1)$$

where $C$ is the non-terminal node which immediately dominates $A$, $a_1$ is the part-of-speech associated with the leftmost word of constituent $A$, and $a_0$ and $a_2$ are the parts-of-speech of the words to the left and to the right of $a_1$, respectively. See Magerman and Marcus 1991 [12] for a more detailed description of the CFG with CSP model.

---

*Special thanks to Jerry Hobbs and Bob Moore at SRI for providing access to their computers, and to Salim Roukos, Peter Brown, and Vincent and Steven Della Pietra at IBM for their instructive lessons on probabilistic modelling of natural language.

[1]$\mathcal{P}$earl ≡ probabilistic Earley-style parser ($\mathcal{P}$-Earl). $\mathcal{P}$icky ≡ probabilistic CKY-like parser ($\mathcal{P}$-CKY).

[2]Some familiarity with chart parsing terminology is assumed in this paper. For terminological definitions, see [9], [10], [11], or [17].

[3]Sections 2 and 3, the descriptions of the probabilistic models used in $\mathcal{P}$icky and the $\mathcal{P}$icky algorithm, are similar in content to the corresponding sections of Magerman and Weir[13]. The experimental results and discussions which follow in sections 4-6 are original.

## 3. The Parsing Algorithm

A probabilistic language model, such as the aforementioned CFG with CSP model, provides a metric for evaluating the likelihood of a parse tree. However, while it may *suggest* a method for evaluating partial parse trees, a language model alone does not dictate the search strategy for determining the most likely analysis of an input. Since exhaustive search of the space of parse trees produced by a natural language grammar is generally not feasible, a parsing model can best take advantage of a probabilistic language model by incorporating it into a parser which probabilistically models the parsing process. $\mathcal{P}$icky attempts to model the chart parsing process for context-free grammars using probabilistic prediction.

$\mathcal{P}$icky parses sentences in three phases: *covered left-corner* phase (I), *covered bidirectional* phase (II), and *tree completion* phase (III). Each phase uses a different method for proposing edges to be introduced to the parse chart. The first phase, covered left-corner, uses probabilistic prediction based on the left-corner word of the left-most daughter of a constituent to propose edges. The covered bidirectional phase also uses probabilistic prediction, but it allows prediction to occur from the left-corner word of any daughter of a constituent, and parses that constituent outward (bidirectionally) from that daughter. These phases are referred to as "covered" because, during these phases, the parsing mechanism proposes only edges that have non-zero probability according to the prediction model, i.e. that have been covered by the training process. The final phase, tree completion, is essentially an exhaustive search of all interpretations of the input according to the grammar. However, the search proceeds in best-first order, according to the measures provided by the language model. This phase is used only when the probabilistic prediction model fails to propose the edges necessary to complete a parse of the sentence.

The following sections will present and motivate the prediction techniques used by the algorithm, and will then describe how they are implemented in each phase.

### 3.1. Probabilistic Prediction

Probabilistic prediction is a general method for using probabilistic information extracted from a parsed corpus to estimate the likelihood that predicting an edge at a certain point in the chart will lead to a correct analysis of the sentence. The $\mathcal{P}$icky algorithm is not dependent on the specific probabilistic prediction model used. The model used in the implementation, which is similar to the probabilistic language model, will be described.[4]

The prediction model used in the implementation of $\mathcal{P}$icky estimates the probability that an edge proposed at a point in the chart will lead to a correct parse to be:

$$\mathcal{P}(A \to \alpha B \beta | a_0 a_1 a_2), \quad (2)$$

where $a_1$ is the part-of-speech of the left-corner word of $B$, $a_0$ is the part-of-speech of the word to the left of $a_1$, and $a_2$ is the part-of-speech of the word to the right of $a_1$.

To illustrate how this model is used, consider the sentence

$$\text{The cow raced past the barn.} \quad (3)$$

The word "cow" in the word sequence "the cow raced" predicts **NP → det n**, but not **NP → det n PP**, since PP is unlikely to generate a verb, based on training material.[5] Assuming the prediction model is well trained, it *will* propose the interpretation of "raced" as the beginning of a participial phrase modifying "the cow," as in

$$\text{The cow raced past the barn mooed.} \quad (4)$$

However, the interpretation of "raced" as a past participle will receive a low probability estimate relative to the verb interpretation, since the prediction model only considers local context.

---

[4]It is not necessary for the prediction model to be the same as the language model used to evaluate complete analyses. However, it is helpful if this is the case, so that the probability estimates of incomplete edges will be consistent with the probability estimates of completed constituents.

[5]Throughout this discussion, we will describe the prediction process using words as the predictors of edges. In the implementation, due to sparse data concerns, only parts-of-speech are used to predict edges. Given more robust estimation techniques, a probabilistic prediction model conditioned on word sequences is likely to perform as well or better.

The process of probabilistic prediction is analogous to that of a human parser recognizing predictive lexical items or sequences in a sentence and using these hints to restrict the search for the correct analysis of the sentence. For instance, a sentence beginning with a *wh*-word and auxiliary inversion is very likely to be a question, and trying to interpret it as an assertion is wasteful. If a verb is generally ditransitive, one should look for two objects to that verb instead of one or none. Using probabilistic prediction, sentences whose interpretations are highly predictable based on the trained parsing model can be analyzed with little wasted effort, generating sometimes no more than ten spurious constituents for sentences which contain between 30 and 40 constituents! Also, in some of these cases every predicted rule results in a completed constituent, indicating that the model made *no* incorrect predictions and was led astray only by genuine ambiguities in parts of the sentence.

## 3.2. Exhaustive Prediction

When probabilistic prediction fails to generate the edges necessary to complete a parse of the sentence, exhaustive prediction uses the edges which have been generated in earlier phases to predict new edges which might combine with them to produce a complete parse. Exhaustive prediction is a combination of two existing types of prediction, "over-the-top" prediction [11] and top-down filtering.

Over-the-top prediction is applied to complete edges. A completed edge $A \rightarrow \alpha$ will predict all edges of the form $B \rightarrow \beta A \gamma$.[6]

Top-down filtering is used to predict edges in order to complete incomplete edges. An edge of the form $A \rightarrow \alpha B_0 B_1 B_2 \beta$, where a $B_1$ has been recognized, will predict edges of the form $B_0 \rightarrow \gamma$ before $B_1$ and edges of the form $B_2 \rightarrow \delta$ after $B_1$.

## 3.3. Bidirectional Parsing

The only difference between phases I and II is that phase II allows bidirectional parsing. Bidirectional parsing is a technique for initiating the parsing of a constituent from any point in that constituent. Chart parsing algorithms generally process constituents from left-to-right. For instance, given a grammar rule

$$A \rightarrow B_1 B_2 \cdots B_n, \qquad (5)$$

---
[6] In the implementation of $\mathcal{P}$icky, over-the-top prediction for $A \rightarrow \alpha$ will only predict edges of the form $B \rightarrow A\gamma$. This limitation on over-the-top prediction is due to the expensive bookkeeping involved in bidirectional parsing. See the section on bidirectional parsing for more details.

a parser generally would attempt to recognize a $B_1$, then search for a $B_2$ following it, and so on. Bidirectional parsing recognizes an $A$ by looking for any $B_i$. Once a $B_i$ has been parsed, a bidirectional parser looks for a $B_{i-1}$ to the left of the $B_i$, a $B_{i+1}$ to the right, and so on.

Bidirectional parsing is generally an inefficient technique, since it allows duplicate edges to be introduced into the chart. As an example, consider a context-free rule NP $\rightarrow$ DET N, and assume that there is a determiner followed by a noun in the sentence being parsed. Using bidirectional parsing, this NP rule can be predicted both by the determiner and by the noun. The edge predicted by the determiner will look to the right for a noun, find one, and introduce a new edge consisting of a completed NP. The edge predicted by the noun will look to the left for a determiner, find one, and also introduce a new edge consisting of a completed NP. Both of these NPs represent identical parse trees, and are thus redundant. If the algorithm permits both edges to be inserted into the chart, then an edge XP $\rightarrow \alpha$ NP $\beta$ will be advanced by both NPs, creating two copies of every XP edge. These duplicate XP edges can themselves be used in other rules, and so on.

To avoid this propagation of redundant edges, the parser must ensure that no duplicate edges are introduced into the chart. $\mathcal{P}$icky does this simply by verifying every time an edge is added that the edge is not already in the chart.

Although eliminating redundant edges prevents excessive inefficiency, bidirectional parsing may still perform more work than traditional left-to-right parsing. In the previous example, three edges are introduced into the chart to parse the NP $\rightarrow$ DET N edge. A left-to-right parser would only introduce two edges, one when the determiner is recognized, and another when the noun is recognized.

The benefit of bidirectional parsing can be seen when probabilistic prediction is introduced into the parser. Frequently, the syntactic structure of a constituent is not determined by its left-corner word. For instance, in the sequence V NP PP, the prepositional phrase PP can modify either the noun phrase NP or the entire verb phrase V NP. These two interpretations require different VP rules to be predicted, but the decision about which rule to use depends on more than just the verb. The correct rule may best be predicted by knowing the preposition used in the PP. Using probabilistic prediction, the decision is made by pursuing the rule which has the highest probability according to the prediction model. This rule is then parsed bidirectionally. If this rule is in fact the correct rule to analyze the constituent, then no other predictions will be made for that constituent, and there will be no more edges produced than in left-to-right parsing. Thus, the only case where bidirectional parsing is less efficient than left-to-right parsing is when the prediction model fails to capture the elements of context of the sentence which determine its correct interpretation.

### 3.4. The Three Phases of $\mathcal{P}$icky

**Covered Left-Corner** The first phase uses probabilistic prediction based on the part-of-speech sequences from the input sentence to predict all grammar rules which have a non-zero probability of being dominated by that trigam (based on the training corpus), i.e.

$$\mathcal{P}(A \to B\delta | a_0 a_1 a_2) > 0 \tag{6}$$

where $a_1$ is the part-of-speech of the left-corner word of $B$. In this phase, the only exception to the probabilistic prediction is that any rule which can immediately dominate the preterminal category of any word in the sentence is also predicted, regardless of its probability. This type of prediction is referred to as *exhaustive prediction*. All of the predicted rules are processed using a standard best-first agenda processing algorithm, where the highest scoring edge in the chart is advanced.

**Covered Bidirectional** If an $S$ spanning the entire word string is not recognized by the end of the first phase, the covered bidirectional phase continues the parsing process. Using the chart generated by the first phase, rules are predicted not only by the trigram centered at the left-corner word of the rule, but by the trigram centered at the left-corner word of any of the children of that rule, i.e.

$$\mathcal{P}(A \to \alpha B \delta | b_0 b_1 b_2) > 0. \tag{7}$$

where $b_1$ is the part-of-speech associated with the leftmost word of constituent $B$. This phase introduces incomplete theories into the chart which need to be expanded to the left and to the right, as described in the bidirectional parsing section above.

**Tree Completion** If the bidirectional processing fails to produce a successful parse, then it is assumed that there is some part of the input sentence which is not covered well by the training material. In the final phase, exhaustive prediction is performed on all complete theories which were introduced in the previous phases but which are not predicted by the trigrams beneath them (i.e. $\mathcal{P}(\text{rule} \mid \text{trigram}) = 0$).

In this phase, edges are only predicted by their left-corner word. As mentioned previously, bidirectional parsing can be inefficient when the prediction model is inaccurate. Since all edges which the prediction model assigns non-zero probability have already been predicted, the model can no longer provide any information for future predictions. Thus, bidirectional parsing in this phase is very likely to be inefficient. Edges already in the chart will be parsed bidirectionally, since they were predicted by the model, but all new edges will be predicted by the left-corner word only.

Since it is already known that the prediction model will assign a zero probability to these rules, these predictions are instead scored based on the number of words spanned by the subtree which predicted them. Thus, this phase processes longer theories by introducing rules which can advance them. Each new theory which is proposed by the parsing process is exhaustively predicted for, using the length-based scoring model.

The final phase is used only when a sentence is so far outside of the scope of the training material that none of the previous phases are able to process it. This phase of the algorithm exhibits the worst-case exponential behavior that is found in chart parsers which do not use node packing. Since the probabilistic model is no longer useful in this phase, the parser is forced to propose an enormous number of theories. The expectation (or hope) is that one of the theories which spans most of the sentence will be completed by this final process. Depending on the size of the grammar used, it may be unfeasible to allow the parser to exhaust all possible predicts before deciding an input is ungrammatical. The question of when the parser should give up is an empirical issue which will not be explored here.

**Post-processing: Partial Parsing** Once the final phase has exhausted all predictions made by the grammar, or more likely, once the probability of all edges in the chart falls below a certain threshold, $\mathcal{P}$icky determines the sentence to be ungrammatical. However, since the chart produced by $\mathcal{P}$icky contains all recognized constituents, sorted by probability, the chart can be used to extract partial parses. As implemented, $\mathcal{P}$icky prints out the most probable completed $S$ constituent.

## 4. Why a New Algorithm?

Previous research efforts have produced a wide variety of parsing algorithms for probabilistic and non-probabilistic grammars. One might question the need for a new algorithm to deal with context-sensitive probabilistic models. However, these previous efforts have generally failed to address both efficiency and robustness effectively.

For non-probabilistic grammar models, the CKY algorithm [9] [17] provides efficiency and robustness in polynomial time, $O(Gn^3)$. CKY can be modified to han-dle simple P-CFGs [2] without loss of efficiency. However, with the introduction of context-sensitive probability models, such as the history-based grammar[1] and the CFG with CSP models[12], CKY cannot be modified to accommodate these models without exhibiting exponential behavior in the grammar size $G$. The linear behavior of CKY with respect to grammar size is dependent upon being able to collapse the distinctions among constituents of the same type which span the same part of the sentence. However, when using a context-sensitive probabilistic model, these distinctions are necessary. For instance, in the CFG with CSP model, the part-of-speech sequence generated by a constituent affects the probability of constituents that dominate it. Thus, two constituents which generate different part-of-speech sequences must be considered individually and cannot be collapsed.

Earley's algorithm [6] is even more attractive than CKY in terms of efficiency, but it suffers from the same exponential behavior when applied to context-sensitive probabilistic models. Still, Earley-style prediction improves the average case performance of en exponential chart-parsing algorithm by reducing the size of the search space, as was shown in [12]. However, Earley-style prediction has serious impacts on robust processing of ungrammatical sentences. Once a sentence has been determined to be ungrammatical, Earley-style prediction prevents any new edges from being added to the parse chart. This behavior seriously degrades the robustness of a natural language system using this type of parser.

A few recent works on probabilistic parsing have proposed algorithms and devices for efficient, robust chart parsing. Bobrow[3] and Chitrao[4] introduce agenda-based probabilistic parsing algorithms, although neither describe their algorithms in detail. Both algorithms use a strictly best first search. As both Chitrao and Magerman[12] observe, a best first search penalizes longer and more complex constituents (i.e. constituents which are composed of more edges), resulting in thrashing and loss of efficiency. Chitrao proposes a heuristic penalty based on constituent length to deal with this problem. Magerman avoids thrashing by calculating the score of a parse tree using the geometric mean of the probabilities of the constituents contained in the tree.

Moore[14] discusses techniques for improving the efficiency and robustness of chart parsers for unification grammars, but the ideas are applicable to probabilistic grammars as well. Some of the techniques proposed are well-known ideas, such as compiling $\epsilon$-transitions (null gaps) out of the grammar and heuristically controlling the introduction of predictions.

The $\mathcal{P}$icky parser incorporates what we deem to be the most effective techniques of these previous works into one parsing algorithm. New techniques, such as probabilistic prediction and the multi-phase approach, are introduced where the literature does not provide adequate solutions. $\mathcal{P}$icky combines the standard chart parsing data structures with existing bottom-up and top-down parsing operations, and includes a probabilistic version of top-down filtering and over-the-top prediction. $\mathcal{P}$icky also incorporates a limited form of bi-directional parsing in a way which avoids its computationally expensive side-effects. It uses an agenda processing control mechanism with the scoring heuristics of $\mathcal{P}$earl.

With the exception of probabilistic prediction, most of the ideas in this work individually are not original to the parsing technology literature. However, the combination of these ideas provides robustness without sacrificing efficiency, and efficiency without losing accuracy.

## 5. Results of Experiments

The $\mathcal{P}$icky parser was tested on 3 sets of 100 sentences which were held out from the rest of the corpus during training. The training corpus consisted of 982 sentences which were parsed using the same grammar that $\mathcal{P}$icky used. The training and test corpora are samples from the MIT's Voyager direction-finding system.[7] Using $\mathcal{P}$icky's grammar, these test sentences generate, on average, over 100 parses per sentence, with some sentences generated over 1,000 parses.

The purpose of these experiments is to explore the impact of varying of $\mathcal{P}$icky's parsing algorithm on parsing accuracy, efficiency, and robustness. For these experiments, we varied three attributes of the parser: the phases used by parser, the maximum number of edges the parser can produce before failure, and the minimum probability parse acceptable.

In the following analysis, the accuracy rate represents the percentage of the test sentences for which the highest probability parse generated by the parser is identical to the "correct" parse tree indicated in the parsed test corpus.[8]

Efficiency is measured by two ratios, the prediction ratio and the completion ratio. The *prediction ratio* is defined as the ratio of number of predictions made by the parser during the parse of a sentence to the number of constituents necessary for a correct parse. The *completion ratio* is the ratio of the number of completed edges to the number of predictions during the parse of sentence.

---

[7]Special thanks to Victor Zue at MIT for the use of the speech data from MIT's Voyager system.

[8]There are two exceptions to this accuracy measure. If the parser generates a plausible parse for a sentences which has multiple plausible interpretations, the parse is considered correct. Also, if the parser generates a correct parse, but the parsed test corpus contains an incorrect parse (i.e. if there is an error in the answer key), the parse is considered correct.

Robustness cannot be measured directly by these experiments, since there are few ungrammatical sentences and there is no implemented method for interpreting the well-formed substring table when a parse fails. However, for each configuration of the parser, we will explore the expected behavior of the parser in the face of ungrammatical input.

Since $\mathcal{P}$icky has the power of a pure bottom-up parser, it would be useful to compare its performance and efficiency to that of a probabilistic bottom-up parser. However, an implementation of a probabilistic bottom-up parser using the same grammar produces on average over 1000 constituents for each sentence, generating over 15,000 edges without generating a parse at all! This supports our claim that exhaustive CKY-like parsing algorithms are not feasible when probabilistic models are applied to them.

### 5.1. Control Configuration

The control for our experiments is the configuration of $\mathcal{P}$icky with all three phases and with a maximum edge count of 15,000. Using this configuration, $\mathcal{P}$icky parsed the 3 test sets with an 89.3% accuracy rate. This is a slight improvement over $\mathcal{P}$earl's 87.5% accuracy rate reported in [12].

Recall that we will measure the efficiency of a parser configuration by its prediction ratio and completion ratio on the test sentences. A perfect prediction ratio is 1:1, i.e. every edge predicted is used in the eventual parse. However, since there is ambiguity in the input sentences, a 1:1 prediction ratio is not likely to be achieved. $\mathcal{P}$icky's prediction ratio is approximately than 4.3:1, and its ratio of predicted edges to completed edges is nearly 1.3:1. Thus, although the prediction ratio is not perfect, on average for every edge that is predicted more than one completed constituent results.

This is the most robust configuration of $\mathcal{P}$icky which will be attempted in our experiments, since it includes bidirectional parsing (phase II) and allows so many edges to be created. Although there was not a sufficient number or variety of ungrammatical sentences to explore the robustness of this configuration further, one interesting example did occur in the test sets. The sentence
How do I how do I get to MIT?
is an ungrammatical but interpretable sentence which begins with a restart. The $\mathcal{P}$earl parser would have generated no analysis for the latter part of the sentence and the corresponding sections of the chart would be empty. Using bidirectional probabilistic prediction, $\mathcal{P}$icky produced a correct partial interpretation of the last 6 words of the sentence, "how do I get to MIT?" One sentence does not make for conclusive evidence, but it represents the type of performance which is expected from the $\mathcal{P}$icky algorithm.

### 5.2. Phases vs. Efficiency

Each of $\mathcal{P}$icky's three phases has a distinct role in the parsing process. Phase I tries to parse the sentences which are most standard, i.e. most consistent with the training material. Phase II uses bidirectional parsing to try to complete the parses for sentences which are nearly completely parsed by Phase I. Phase III uses a simplistic heuristic to glue together constituents generated by phases I and II. Phase III is obviously inefficient, since it is by definition processing atypical sentences. Phase II is also inefficient because of the bidirectional predictions added in this phase. But phase II also amplifies the inefficiency of phase III, since the bidirectional predictions added in phase II are processed further in phase III.

| Phases | Pred. Ratio | Comp. Ratio | Coverage | %Error |
|---|---|---|---|---|
| I | 1.95 | 1.02 | 75.7% | 2.3% |
| I,II | 2.15 | 0.94 | 77.0% | 2.3% |
| II | 2.44 | 0.86 | 77.3% | 2.0% |
| I,III | 4.01 | 1.44 | 88.3% | 11.7% |
| III | 4.29 | 1.40 | 88.7% | 11.3% |
| I,II,III | 4.30 | 1.28 | 89.3% | 10.7% |
| II,III | 4.59 | 1.24 | 89.7% | 10.3% |

Table 1: Prediction and Completion Ratios and accuracy statistics for $\mathcal{P}$icky configured with different subsets of $\mathcal{P}$icky's three phases.

In Table 1, we see the efficiency and accuracy of $\mathcal{P}$icky using different subsets of the parser's phases. Using the control parser (phases I, II, and II), the parser has a 4.3:1 prediction ratio and a 1.3:1 completion ratio.

By omitting phase III, we eliminate nearly half of the predictions and half the completed edges, resulting in a 2.15:1 prediction ratio. But this efficiency comes at the cost of coverage, which will be discussed in the next section.

By omitting phase II, we observe a slight reduction in predictions, but an increase in completed edges. This behavior results from the elimination of the bidirectional predictions, which tend to generate duplicate edges. Note that this configuration, while slightly more efficient, is less robust in processing ungrammatical input.

## 5.3. Phases vs. Accuracy

For some natural language applications, such as a natural language interface to a nuclear reactor or to a computer operating system, it is imperative for the user to have confidence in the parses generated by the parser. $\mathcal{P}$icky has a relatively high parsing accuracy rate of nearly 90%; however, 10% error is far too high for fault-intolerant applications.

| Phase | No. | Accuracy | Coverage | %Error |
|---|---|---|---|---|
| I + II | 238 | 97% | 77% | 3% |
| III | 62 | 60% | 12% | 40% |
| Overall | 300 | 89.3% | 89.3% | 10.7% |

Table 2: $\mathcal{P}$icky's parsing accuracy, categorized by the phase which the parser reached in processing the test sentences.

Consider the data in Table 2. While the parser has an overall accuracy rate of 89.3%, it is far more accurate on sentences which are parsed by phases I and II, at 97%. Note that 238 of the 300 sentences, or 79%, of the test sentences are parsed in these two phases. Thus, by eliminating phase III, the percent error can be reduced to 3%, while maintaining 77% coverage. An alternative to eliminating phase III is to replace the length-based heuristic of this phase with a secondary probabilistic model of the difficult sentences in this domain. This secondary model might be trained on a set of sentences which cannot be parsed in phases I and II.

## 5.4. Edge Count vs. Accuracy

In the original implementation of the $\mathcal{P}$icky algorithm, we intended to allow the parser to generate edges until it found a complete interpretation or exhausted all possible predictions. However, for some ungrammatical sentences, the parser generates tens of thousands of edges without terminating. To limit the processing time for the experiments, we implemented a maximum edge count which was sufficiently large so that all grammatical sentences in the test corpus would be parsed. All of the grammatical test sentences generated a parse before producing 15,000 edges. However, some sentences produced thousands of edges only to generate an incorrect parse. In fact, it seemed likely that there might be a correlation between very high edge counts and incorrect parses. We tested this hypothesis by varying the maximum edge count.

In Table 3, we see an increase in efficiency and a decrease in accuracy as we reduce the maximum number of edges the parser will generate before declaring a sentence ungrammatical. By reducing the maximum edge count by a factor of 50, from 15,000 to 300, we can nearly cut in half the number of predicts and edges generated by the parser. And while this causes the accuracy rate to fall from 89.3% to 79.3%, it also results in a significant decrease in error rate, down to 2.7%. By decreasing the maximum edge count down to 150, the error rate can be reduced to 1.7%.

| Maximum Edge Count | Pred. Ratio | Comp. Ratio | Coverage | %Error |
|---|---|---|---|---|
| 15,000 | 4.30 | 1.35 | 89.3% | 10.7% |
| 1,000 | 3.69 | 0.93 | 83.3% | 7.0% |
| 500 | 3.08 | 0.82 | 80.3% | 5.3% |
| 300 | 2.50 | 0.86 | 79.3% | 2.7% |
| 150 | 1.95 | 0.92 | 66.0% | 1.7% |
| 100 | 1.60 | 0.84 | 43.7% | 1.7% |

Table 3: Prediction and Completion Ratios and accuracy statistics for $\mathcal{P}$icky configured with different maximum edge count.

## 5.5. Probability vs. Accuracy

Since a probability represents the likelihood of an interpretation, it is not unreasonable to expect the probability of a parse tree to be correlated with the accuracy of the parse. However, based on the probabilities associated with the "correct" parse trees of the test sentences, there appears to be no such correlation. Many of the test sentences had correct parses with very low probabilities ($10^{-10}$), while others had much higher probabilities ($10^{-2}$). And the probabilities associated with incorrect parses were not distinguishable from the probabilities of correct parses.

The failure to find a correlation between probability and accuracy in this experiment does not prove conclusively that no such correlation exists. Admittedly, the training corpus used for all of these experiments is far smaller than one would hope to estimate the CFG with CSP model parameters. Thus, while the model is trained well enough to steer the parsing search, it may not be sufficiently trained to provide meaningful probability values.

## 6. Conclusions

There are many different applications of natural language parsing, and each application has a different cost threshold for efficiency, robustness, and accuracy. The $\mathcal{P}$icky algorithm introduces a framework for integrating these thresholds into the configuration of the parser in order to maximize the effectiveness of the parser for the task at hand. An application which requires a high degree of accuracy would omit the Tree Completion phase of the parser. A real-time application would limit the number of edges generated by the parser, likely at the cost of accuracy. An application which is robust to errors but requires efficient processing of input would omit the Covered Bidirectional phase.

The $\mathcal{P}$icky parsing algorithm illustrates how probabilistic modelling of natural language can be used to improve the efficiency, robustness, and accuracy of natural language understanding tools.

## REFERENCES

1. Black, E., Jelinek, F., Lafferty, J., Magerman, D. M., Mercer, R. and Roukos, S. 1992. Towards History-based Grammars: Using Richer Models of Context in Probabilistic Parsing. In Proceedings of the February 1992 DARPA Speech and Natural Language Workshop. Arden House, NY.
2. Brown, P., Jelinek, F., and Mercer, R. 1991. Basic Method of Probabilistic Context-free Grammars. IBM Internal Report. Yorktown Heights, NY.
3. Bobrow, R. J. 1991. Statistical Agenda Parsing. In Proceedings of the February 1991 DARPA Speech and Natural Language Workshop. Asilomar, California.
4. Chitrao, M. and Grishman, R. 1990. Statistical Parsing of Messages. In Proceedings of the June 1990 DARPA Speech and Natural Language Workshop. Hidden Valley, Pennsylvania.
5. Church, K. 1988. A Stochastic Parts Program and Noun Phrase Parser for Unrestricted Text. In Proceedings of the Second Conference on Applied Natural Language Processing. Austin, Texas.
6. Earley, J. 1970. An Efficient Context-Free Parsing Algorithm. *Communications of the ACM* Vol. 13, No. 2, pp. 94-102.
7. Gale, W. A. and Church, K. 1990. Poor Estimates of Context are Worse than None. In Proceedings of the June 1990 DARPA Speech and Natural Language Workshop. Hidden Valley, Pennsylvania.
8. Jelinek, F. 1985. Self-organizing Language Modeling for Speech Recognition. IBM Report.
9. Kasami, T. 1965. An Efficient Recognition and Syntax Algorithm for Context-Free Languages. Scientific Report AFCRL-65-758, Air Force Cambridge Research Laboratory. Bedford, Massachusetts.
10. Kay, M. 1980. Algorithm Schemata and Data Structures in Syntactic Processing. *CSL-80-12*, October 1980.
11. Kimball, J. 1973. Principles of Surface Structure Parsing in Natural Language. *Cognition*, 2.15-47.
12. Magerman, D. M. and Marcus, M. P. 1991. Pearl: A Probabilistic Chart Parser. In Proceedings of the European ACL Conference, March 1991. Berlin, Germany.
13. Magerman, D. M. and Weir, C. 1992. Probabilistic Prediction and $\mathcal{P}$icky Chart Parsing. In Proceedings of the February 1992 DARPA Speech and Natural Language Workshop. Arden House, NY.
14. Moore, R. and Dowding, J. 1991. Efficient Bottom-Up Parsing. In Proceedings of the February 1991 DARPA Speech and Natural Language Workshop. Asilomar, California.
15. Sharman, R. A., Jelinek, F., and Mercer, R. 1990. Generating a Grammar for Statistical Training. In Proceedings of the June 1990 DARPA Speech and Natural Language Workshop. Hidden Valley, Pennsylvania.
16. Seneff, Stephanie 1989. TINA. In Proceedings of the August 1989 International Workshop in Parsing Technologies. Pittsburgh, Pennsylvania.
17. Younger, D. H. 1967. Recognition and Parsing of Context-Free Languages in Time $n^3$. *Information and Control* Vol. 10, No. 2, pp. 189-208.